# SOLVING THE EXAM SCHEDULING PROBLEMS IN CENTRAL EXAMS WITH GENETIC ALGORITHMS

Murat DENER
Graduate School of Natural and Applied Sciences, Gazi University, Turkey, muratdener@gazi.edu.tr
https://orcid.org/0000-0001-5746-6141
*M. Hanefi CALP
Software Engineering, OF Technology Faculty, Karadeniz Technical University, Turkey, mhcalp@ktu.edu.tr
https://orcid.org/0000-0001-7991-438X

**Abstract**

*It is the efficient use of resources expected from an exam scheduling application. There are various criteria for efficient use of resources and for all tests to be carried out at minimum cost in the shortest possible time. It is aimed that educational institutions with such criteria successfully carry out central examination organizations. In the study, a two-stage genetic algorithm was developed. In the first stage, the assignment of courses to sessions was carried out. In the second stage, the students who participated in the test session were assigned to examination rooms. Purposes of the study are increasing the number of joint students participating in sessions, using the minimum number of buildings in the same session, and reducing the number of supervisors using the minimum number of classrooms possible. In this study, a general purpose exam scheduling solution for educational institutions was presented. The developed system can be used in different central examinations to create originality. Given the results of the sample application, it is seen that the proposed genetic algorithm gives successful results.*

# MERKEZİ SINAVLARDA YAŞANAN SINAV ÇİZELGELEME PROBLEMLERİNİN GENETİK ALGORİTMALAR İLE ÇÖZÜLMESİ

**Öz**

*Bir sınav çizelgeleme uygulamasından beklenen kaynakların verimli kullanımıdır. Kaynakları verimli kullanabilmek ve en kısa zamanda en az maliyetle bütün sınavların gerçekleştirilmesi için çeşitli kıstaslar vardır. Yapılan çalışma ile bu tür kıstaslara sahip eğitim kurumlarının, merkezi sınav organizasyonlarını başarıyla gerçekleştirmesi amaçlanmıştır. Çalışmada, iki aşamalı genetik algoritma geliştirilmiştir. Birinci aşamada derslerin oturumlara atanması işlemi, ikinci aşamada ise ilgili oturumda sınava katılacak öğrencilerin sınav salonlarına atanması işlemi gerçekleştirilmiştir. Oturumlara katılan ortak öğrenci sayısının arttırılması, aynı oturumda asgari bina kullanımı, mümkün olan en az sayıda sınıf-sıra kullanılarak gözetmen sayısının azaltılması yapılan çalışmanın amaçlarını oluşturmaktadır. Yapılan bu çalışmada eğitim kurumlarına yönelik genel amaçlı sınav çizelgeleme çözümü sunulmuştur. Geliştirilen sistem farklı merkezi sınavlar içinde kullanılabilmesiyle özgünlük oluşturmaktadır. Örnek uygulama sonuçlarına bakıldığında önerilen genetik algoritmanın başarılı sonuçlar verdiği görülmektedir.*

**Anahtar Kelimeler: Merkezi Sınavlar, Genetik Algoritma, Sınav Planlama Problemi, Çözme**

## 1. Introduction

Scheduling is the process of assigning a certain number of activities to a limited number of resources and time periods to provide certain conditions. Scheduling problem is defined as "allocating resources in time space, under certain constraints, as much as possible to reach the specified targets" [1]. It is aimed to plan the appropriate time slots for tasks using limited resources in scheduling problems. Numerous studies have been conducted to solve the scheduling problems that are of interest to over a decade of researchers. There are applications for solving many problems in the field such as schedule of the operating rooms with nurses and doctors' watches in hospitals, exam calendars with training schedule of workshops at training institutions, scheduling of airplane landing and departure times at airports, planning sporting events.

A sub-type of scheduling problems and the problem of scheduling problem in the class of scheduling problem are needed to be solved intensively in educational institutions today. The test scheduling problem is an optimization problem that is accepted in the NP-hard class because of the constraints and the complexity of the components. In such problems, it is not always possible to reach a hundred percent solution, and when the best solution that can be achieved is obtained, the solution is terminated and the solution found is accepted as the best solution. Depending on the increase in the number of departments, programs, teachers and students in the educational institutions, the level of complexity of the central examination programs is increasing.

This situation makes it inevitable to prepare the central examination programs with a systematic approach. In addition, manual preparation of the central examination programs, inexperience of the staff, carelessness or lack of sight cause many mistakes on the programs. The difficulty of resolving the problem with classical methods also leads to the failure to meet the desired conditions. Also, preparing the charts in this way causes serious loss of time and work power. Therefore, in this study, a general-purpose solution for the central examination organization of educational institutions is presented by using genetic algorithm technique which is a heuristic approach. Heuristic approachs can be said that they are effective for most problems. In realized studies, it seen that the adoption of meta-heuristics such as particle swarm optimization, antlion optimization, simulated annealing, bee colony algorithm, tabu search and genetic algorithms has produce better results than classical dispatching or greedy heuristic algorithms [2–5]. In this context, by means of a two-stage genetic algorithm developed within the scope of solution of the central examination problem, the scheduling problem of examinations to be done by the resources at hand is optimized.

In the study, the assignment of the courses to the sessions and the assignment of the students who will participate in the test in the relevant sessions to the examination rooms were carried out. It is aimed to minimize the manpower and time spent in the exam charts created using the developed software with the classical methods and to meet the requirements and requests at the optimum level in the preparation of the charts. In the literature, especially, the lack of the approaches of artificial intelligence in the central exam scheduling are an important element in terms of the originality of the study and its contribution to this subject. The other parts of the study are as follows: In Chapter 2 is given information about the problem of exam scheduling and the problem is solved by genetic algorithm. The design of the algorithm proposed in Chapter 3 is described in detail. The results of the sample application are presented in Chapter 4 and the overall results of the study are given in Chapter 5.

**1.1. Related Studies**

Due to differences in the requirements and constraints of each test organization, no inclusive solution or formulation can be obtained for the problem of exam scheduling. For this reason, the problem of exam scheduling is one of the topics that researchers have worked extensively on. This diversity brings with it different approaches to solve the problem. Bulut et al. [6] have identified a new seating plan with a genetic algorithm for centralized exams. Their aim is to reduce the probability of copying at the central examinations as much as possible. Genetic Algorithms are used in the software to prevent people who are likely to know each other from testing in the same examination rooms. In the application, only name, surname, school, address information of the students are used. Software has been realized in this direction, considering that people with the same surname, people sitting on the same street, and people reading at the same school may be relatives or friends. Yaldır et al. [7] used the Evolutionary Calculation method to automate the application of the test calendars prepared manually at certain periods in the universities. The studies consist of two parts. In the first part, the data is collected from the related units on the web environment and in the second part, the developed desktop application is executed and desired results are obtained. The study includes examinations, supervisors who will take part in these exams, classrooms where examinations will take place, and data on which students will take the exam in which classrooms. The examinations are randomly placed and the lists are created so that there is no overlap. Özdağ et al. [8] developed a tree graph model suitable for the Ant Colony Algorithm of the course distribution optimization problem by assigning grades in the direction of the day and time demands of the teaching members and restricting the lectures and teaching members on the model. İlkuçar [9] developed an application for the optimization of the scheduling problem of examinations such as midterm and final examinations in universities. With the developed application software, exam - room layout, supervisory layout and conflicts were removed, balanced distribution of tasks was achieved and requests were made as much as possible. The software was used in the distribution of midterm and final exams at Mehmet Akif Ersoy University Vocational High School. Rozaimee et al. aimed to present a University's Final Exam Timetable Allocation using genetic algorithm method. They utilized the method they suggested in order that provide an optimal solution of solving timetable by using a real data from UniSZA's academic department. As result, proposed method was reducing the time taken in timetable allocation [10].

Fraiwan and Al-Qahtani described the efforts of the Arab East College for High Education in Saudi Arabia in scheduling exams in the minimum number of conflicts. They proposed a two-stage solution approach in the study. The first phase is a greedy algorithm and the next is a genetic algorithm. The two algorithms work in synchronous to create the best exam timetable. Experimental results indicated that reduced the number of conflicts, exam days, and the required venues [11]. Moreira suggested a solution method to the problem of automatic timetables for the exams and utilized the genetic algorithm technique for solving the exam problems. That is, the method of solution is a meta-heuristics that includes a genetic algorithm. The suggested model is used for the building of exam timetables in institutions of higher education. The results obtained from real and complex experiments indicated the success of the proposed model. As result, the model is satisfactory, becasue the exam timetabling is fulfills the realized regulations [12]. Mandal and Kahar organized the exams based on graph heuristics. They scheduled only selected exams (partial exams) and proposed a model using great deluge algorithm. In addition, they applied the proposed method on the Toronto benchmark datasets. Experimental results showed that the proposed method better than traditional great deluge algorithm, perform successful solution for all samples and better reported results [13].

Gershil et al. [14] have developed a software program that prepares and optimizes the curriculum using genetic algorithms in their study. The software program developed to provide more effective education and training uses the efficiency of education and teaching and the weight of the course as criteria. Although there are many studies in the literature about solving the problem of exam scheduling by genetic algorithm [15-17], there are not many studies for central examinations. In addition, there are different studies [18-24] that seek solutions to the problem of exam scheduling with different techniques such as fuzzy logic and data mining. However, these studies do not cover the central examinations again. It is evaluated that the study done in this respect will contribute to the literature.

## 2. Solving the Exam Scheduling Problems in Central Exams with Genetic Algorithm

### 2.1. Exam Scheduling Problem

Scheduling problems aim at allocating resources with certain constraints to users, based on efficiency. The problem of exam scheduling is the problem of assigning a set of tests to a specified number of given time intervals and under specified constraints [25]. The goal in solving the problem is to plan exams for the least amount of time without overlapping and to try to maximize priorities such as increasing the time interval between management and trainers and the two consecutive exams of the students as much as possible when the number of time periods is fixed [26]. There are limitations to be considered in solving the problem.

Session / course assignments, class / hall capacities, courses for students to take exams, the amount of time that must be left between exams, the maximum number of exams that a student can enter per day, the type of exam room that should be taken according to the exam, the beginning and end times of the exam, the specific preferences of the instructors and exam supervisors, and the start and end dates of the exam period are some of these constraints.

The mentioned constraints are treated in two groups as hard and soft constraints. The constraints that determine the solution space and are not to be violated are hard; the constraints that can be violated with a certain punishment and contribute to the objective function are defined as soft constraints [27].

Examples of soft constraints that students do not have to come to the same time, such as not meeting the capacity of the room to be tested, hard constraints that can not be violated, special preferences of supervisors, late examinations not being put in late hours are not obligatory but are tried to be fulfilled as much as possible. The extent to which soft constraints can be met while preparing the exam charts will increase the success in terms of satisfying the requirements of the charts. The hard constraints to be considered in the study are: Multiple occurrences of the same course in the same session, use of a class in a session, once. Soft constraints that are taken into account in the study are: The number of common students in the same session is too much, the number of different students in the same session is little, the number of vacancies is minimum, the total number of supervisors is minimum, and the number of different buildings is minimum.

### 2.2. Solving the Problem with Genetic Algorithm

The proposed solution consists of two steps. At the first stage, courses are assigned to sessions. In the second stage, the students who will participate in the test session are assigned to the halls. Both stages have benefited from the Genetic Algorithm.

#### 2.2.1. Assigning courses to sessions

The first step in this phase is the determination of the number of sessions required. First of all, the maximum duration of a session must be pre-defined. In addition, the duration of a course must not exceed the maximum session length. Formula 1 was used to determine the number of sessions. The duration of the exam for each course is entered in minutes. Here, all the courses are divided into the maximum session length, which is the sum of the minutes of the examination (Tci) in minutes. The result obtained is rounded up to obtain the number of sessions required. For example, 4 sessions are required in an exam organization where the total duration of examinations for all courses is 570 min and the maximum session duration is 150 min.

$$Number\ of\ Sessions = Round\ up\ [(\sum_{i=1}^{n} Tci) Maximum\ Session\ Length \quad (1)$$

After the number of sessions is determined, the next step is to determine the courses to take place in the exam examination organization. For each course in each department, course code, course name, credits, department code, duration of the exam are held.

*Course Assignment Algorithm:* The flow diagram of the Course Assignment Algorithm is given in Figure 1.

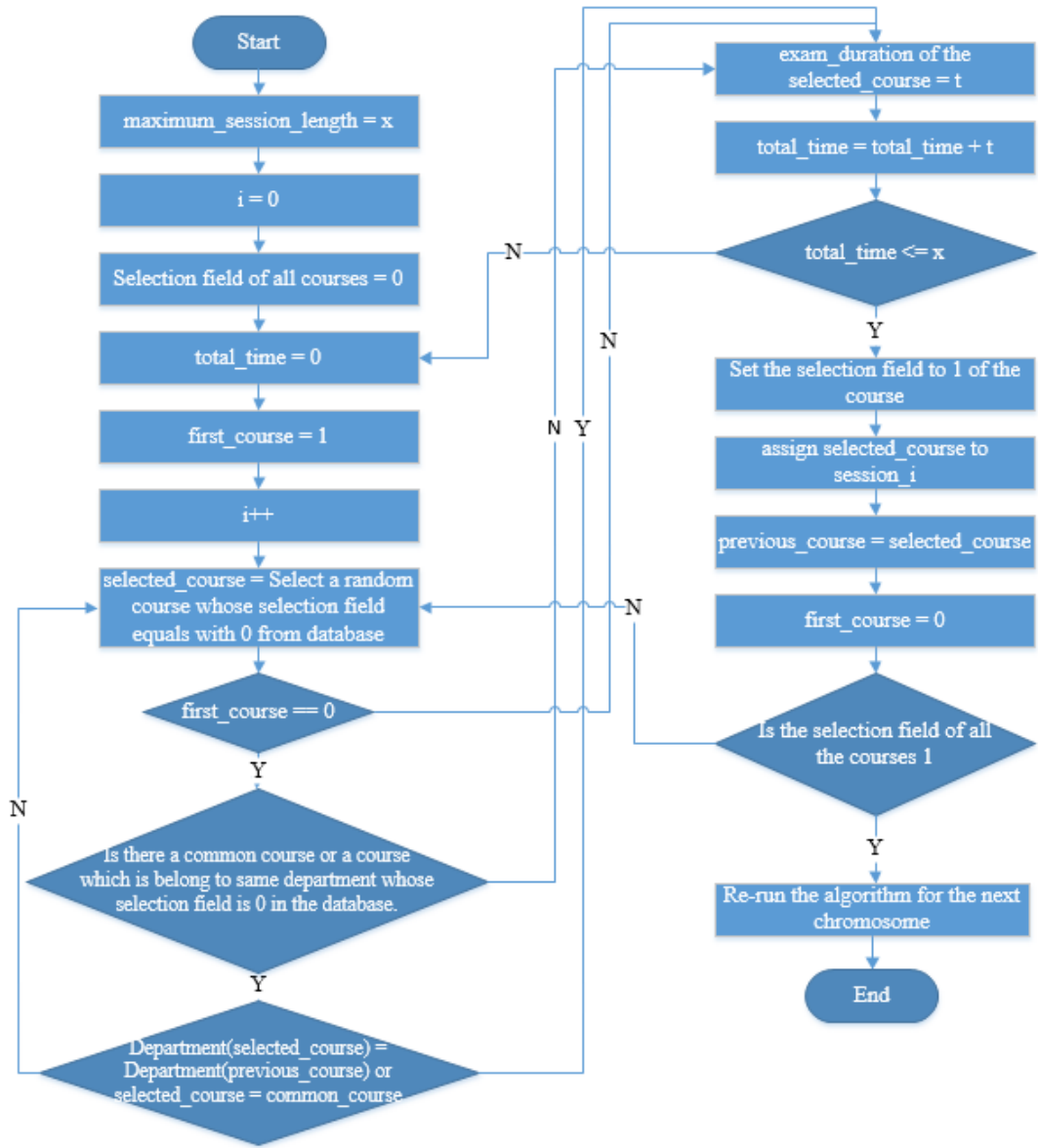

Figure 1. Flowchart of course assignment algorithm

Courses are assigned to relevant sessions with the Course Assignment Algorithm. As shown in Figure 1, a random course is selected from the course database. Exam periods are added for the selected course. When the courses are selected, the previous course is checked to see if the course is in the same department or common course. The reason for this is to ensure that the courses of the different departments of the sessions are at a minimum. At the same time, when the study of the algorithm is approaching the end, all of these courses can be assigned to a session if only the courses of different departments are left in the course database. As the taken into consideration selection area of the courses in each choice, the repetition and loop events are prevented. When the sum of the exam periods of the selected courses equals the maximum session duration, the next session is started. After the courses of all the defined sessions have been assigned, one chromosome finishes and the same process continues for the next chromosome.

*Initial Population:* The Course Assignment algorithm is running until the specified population is completed. Since there are no repetition and loop events in the Course Assignment algorithm, no new control is needed here. The application is run at different numbers for the population size. The size of the population that achieved the best solution in the shortest time was determined as 10. Selected chromosomes are randomly selected from all course alternatives. In Figure 2, the example chromosome structure is shown. The genes shown in Figure 2 refer to the courses in the sessions.

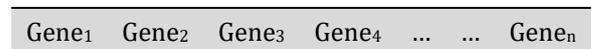

| $Gene_1$ | $Gene_2$ | $Gene_3$ | $Gene_4$ | … | … | $Gene_n$ |

Figure 2. Chromosome structure

*Fitness Function:* The determination of the fitness function is the factor that influences the effectiveness of the genetic algorithm at the most important level. The probability of the appropriateness function, which has not been established in a precisely appropriate manner, can prolong the working period, and even cause the solution to never be reached [28]. Courses are assigned to sessions for each chromosome to determine the fitness function. Subsequently, the total number of Exam Periods, Number of Partner Students and Number of Different Students are obtained for the courses assigned to the sessions. The number of partner students means the number of students who will take the examinations of all the courses in the session. The number of different students means the number of different students who take the examinations of all the courses in the session. The fact that the number of partner students is high makes it possible for students to take lessons for lessons that they have taken, in less sessions. It also prevents students from participating in too many sessions for a few courses. Therefore, the fitness function is obtained by subtracting the number of common students (CS) determined for each session from the number of different students (DS). If the DS value is greater than the CS value, the result is negative. The fitness function is obtained when the values are subtracted from the minimum value in the population. The fitness function is given in formula 2. The i value in the form refers to the number of sessions.

$$FF = \left(\sum_{i=1}^{n} CS_i - DS_i\right) - Min\,(CS_i - DS_i) \quad (2)$$

*Elitism:* Elitism is used for selection in genetic algorithm. If the elite individual is not taken, the best individual of the new generation may be worse than the best individual of the previous generation. As the number of elite individuals increases, the diversity in the solution decreases. Hence, the best individual in a given number must be transferred directly to the new generation without any processing [28]. In this study, two individuals with high scores were transferred to the new population to increase the number of partner students and decrease the number of different students.

*Selection:* The roulette wheel method is used for selection. On the roulette wheel, each individual has a greater chance of being transferred to the new population as the solvency rate increases [28]. The number of common students and the number of different students in the chromosomes in the population are then calculated and then the difference between these values is calculated. Since the difference between chromosomes where the number of different students is higher is negative, all the results are subtracted from the minimum value found in the population. As is known, selecting the Roulette wheel requires that the conformance values of the solutions are not negative. Because, if the probabilities are negative, these solutions have no chance of being selected. In a population of which the majority of the eligibility criteria are negative, new generations may be stuck with certain points. The operation in Formula 3 has been carried out for selection. As shown in formula 3, once the fitness value of each chromosome has been calculated, the value obtained is divided by the sum of the fitness values of all the chromosomes, the result is rounded, and then multiplied by 100.

$$=ROUND(CS_i\text{-}DS_i\text{-}Min\,(CS_i - DS_i)/Total;2)*100 \quad (3)$$

In this way a random number from 0 to 100 is chosen. Since the range to be reached is high, the probability of that combination is high. In this way, an individual selection process takes place.

*Crossover:* Crossover is the exchange of genes between two chromosomes to form two new chromosomes. It is one of the most important operators in genetic algorithm. In the study, multi-point crossing between individuals was applied. Primarily chromosomes are divided into session by sessions. The two black lines in Figure 3 represent courses in one session. When the duration of the courses is equal, the number of courses in the session can create equality. There are 4 sessions in the sample shown. Figure 3 shows the crossover operation.

**Before Crossover**

| Gene$_1$ | Gene$_2$ | Gene$_3$ | Gene$_4$ | Gene$_5$ | Gene$_6$ | Gene$_7$ | Gene$_8$ | Gene$_9$ | ... | ... | Gene$_n$ |
|---|---|---|---|---|---|---|---|---|---|---|---|
| Gene$_1$ | Gene$_2$ | Gene$_3$ | Gene$_4$ | Gene$_5$ | Gene$_6$ | Gene$_7$ | Gene$_8$ | Gene$_9$ | ... | ... | Gene$_n$ |

**After Crossover**

| Gene1 | Gene$_2$ | Gene$_3$ | Gene$_4$ | Gene$_5$ | Gene$_6$ | Gene$_7$ | Gene$_8$ | Gene$_9$ | ... | ... | Gene$_n$ |
|---|---|---|---|---|---|---|---|---|---|---|---|
| Gene1 | Gene$_2$ | Gene$_3$ | Gene$_4$ | Gene$_5$ | Gene$_6$ | Gene$_7$ | Gene$_8$ | Gene$_9$ | ... | ... | Gene$_n$ |

Figure 3. Crossover process

*Mutation:* After the implementation of the crossover operator every generation, there is a series of sequences with the same genetic structure in the succeeding generations. In this case, the more the crossover operator is being applied, the more the diversity in the genes of the children born to the parents is caused. To remove this situation, the exchange (mutation) operator is applied [29]. The purpose of this operator is to replace the genes in the same individual, where each one is a repetition and diversity is reduced. The mutation process performed in the study is given in Fig 4.

**Before Mutation**

| Gene$_1$ | Gene$_2$ | Gene$_3$ | Gene$_4$ | Gene$_5$ | Gene$_6$ | Gene$_7$ | Gene$_8$ | Gene$_9$ | … | … | Gene$_n$ |

**After Mutation**

| Gene$_1$ | Gene$_2$ | **Gene$_8$** | Gene$_4$ | Gene$_5$ | Gene$_6$ | Gene$_7$ | **Gene$_3$** | Gene$_9$ | … | … | Gene$_n$ |

Figure 4. Mutation process

*Repair Operator:* Disappearance of existing information as a result of the processes performed to ensure chromosome diversity, the introduction of extra unwanted information, or the deterioration of chromosome structures may occur [29]. In this part of the study, there is no need for a repair operator in the chromosomes where the mutation process is performed, and the repair operator is needed in the cross chromosomes. The repair operator works as follows. The same course can be more than one in new individuals that occur after the crossover. On the other hand, a course should take place once on a chromosome. Because, a course takes place in a session and is a test once. In order to remove this situation, the same genes are detected, and genes that are never used in that chromosome are randomly assigned. This way, the structure of the chromosome is corrected.

### 2.2.2. Assigning students who are in session to classes

After assigning the courses to the sessions and finding the best result with the genetic algorithm, it is the process of assigning the students who are in these sessions to the classes. For each class, the building name, class name, quota and supervisor requirement are kept.
Class Assignment Algorithm: The flow diagram of the Class Assignment Algorithm is given in Fig 5.

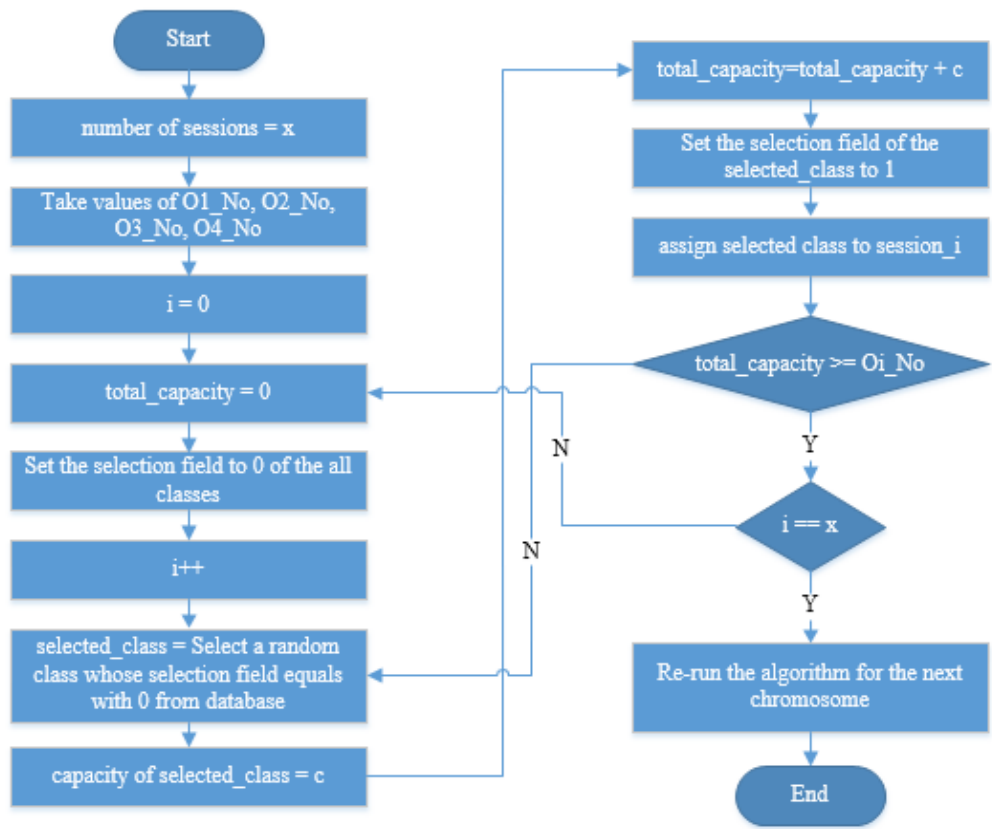

Figure 5. Flowchart for class assignment algorithm

With the Class Assignment Algorithm, the students who are in the relevant sessions are assigned to the classes. As shown in Figure 5, a random class is first selected from the class database. The quotas of the selected class are added. Because the selection area of the classes in the database is checked at each selection, repetition and loop events are prevented. When the sum of the quotas of the selected classes is equal to or greater than the number of students participating in the session, the next session is started. Once the classes belonging to all of the defined sessions have been assigned, a chromosome is running out and the same process continues for the next chromosome.

*Initial Population:* The Class Assignment algorithm runs until the population is complete. There is no need for a new check in this section because the algorithm has been prevented from repeating and looping events. For the population size, the application was run in different numbers and the best size of the population was found to be 10. Random selection is performed among all class alternatives of the selected chromosomes. An exemplary chromosome structure is given in Fig 6. The Genes shown in Figure 6 indicate the classes to be used in the sessions.

| Gene$_1$ | Gene$_2$ | Gene$_3$ | Gene$_4$ | ... | ... | Gene$_n$ |
|---|---|---|---|---|---|---|

Figure 6. Chromosome structure

*Fitness Function:* Classes assigned to sessions for each chromosome are subtracted to determine the fitness function. Then, the number of vacancies in the classroom, the total supervisor requirement, and the number of buildings used in each session are obtained. It is desirable that the number of vacant rows, the total number of supervisors and the number of different buildings are at least equal. First, as shown in Formula 4, for each session (i), the number of vacancies in the classroom (VC), total supervisor requirement (TS) and different building number (DB) values are added. This result is multiplied by 10 since the result in DB is small but effective. The less the result in Formula 4, the more positive this result is. Therefore, the fitness function is determined as given in Formula 5.

$$F = \sum_{i=1}^{n} VCi + TSi + DBi * 10 \quad (4)$$

$$FF = 1 / F \quad (5)$$

*Elitism:* In this section, two highly rated individuals were transferred to the new population to reduce the number of vacancies, reduce the total supervisor requirement to minimum, and reduce the number of different buildings.

*Selection:* The roulette wheel method is applied for selection. After the number of vacancies, the total supervisor requirement and the number of different buildings for the chromosomes in the population are determined, Formula 6 is realized between these values. As shown in formula 6, once the fitness value of each chromosome has been calculated, the value obtained is divided by the sum of the fitness values of all the chromosomes, the result is rounded, and then multiplied by 100.

$$=ROUND(VC\ -TS\ -\ DB*10/\ Total;0)*100 \quad (6)$$

In this way a random number from 0 to the last value is chosen. Since the range to be reached is too high, it is likely that it will arrive. This is how individuals are selected.

*Crossover:* Multi-point crossing is also applied in this section. Figure 7 shows the crossing operation. The two black line ranges shown in Figure 7 refer to the classes to be used in each session. Since the number of students in each session is different, the number of classes to be used can also vary. There are 4 sessions in the sample shown. Then a crossing operation is performed. Figure 7 shows the crossover operation.

*Mutation:* The mutation process performed in this part of the study is shown in Fig 8.

**Before Crossover**

| Gene$_1$ | Gene$_2$ | Gene$_3$ | Gene$_4$ | Gene$_5$ | Gene$_6$ | Gene$_7$ | Gene$_8$ | Gene$_9$ | ... | ... | Gene$_n$ |
|---|---|---|---|---|---|---|---|---|---|---|---|
| Gene$_1$ | Gene$_2$ | Gene$_3$ | Gene$_4$ | Gene$_5$ | Gene$_6$ | Gene$_7$ | Gene$_8$ | Gene$_9$ | ... | ... | Gene$_n$ |

**After Crossover**

| Gene$_1$ | Gene$_2$ | Gene$_3$ | Gene$_4$ | Gene$_5$ | Gene$_6$ | Gene$_7$ | Gene$_8$ | Gene$_9$ | ... | ... | Gene$_n$ |
|---|---|---|---|---|---|---|---|---|---|---|---|
| Gene$_1$ | Gene$_2$ | Gene$_3$ | Gene$_4$ | Gene$_5$ | Gene$_6$ | Gene$_7$ | Gene$_8$ | Gene$_9$ | ... | ... | Gene$_n$ |

Figure 7. Crossover process

**Before Mutation**

| Gene$_1$ | Gene$_2$ | Gene$_3$ | Gene$_4$ | Gene$_5$ | Gene$_6$ | Gene$_7$ | Gene$_8$ | Gene$_9$ | ... | ... | Gene$_n$ |
|---|---|---|---|---|---|---|---|---|---|---|---|

**After Mutation**

| Gene$_1$ | **Gene$_6$** | Gene$_3$ | Gene$_4$ | Gene$_5$ | **Gene$_2$** | Gene$_7$ | Gene$_8$ | Gene$_9$ | ... | ... | Gene$_n$ |
|---|---|---|---|---|---|---|---|---|---|---|---|

Figure 8. Mutation process

*Repair Operator:* In this part of the study, both mutation-treated chromosomes and cross-chromosomes require a repair operator. The repair operator works as follows. New individuals that occur after mutation or crossover may have more than one of the same class in the same session. However, a class can be used once in a session. To remove this situation, genes that are identical for the same session are detected, and instead, genes that are never used for that session of that chromosome are randomly assigned. Another problem is that the number of students to participate in the session is much higher than the quota value because the class and the quota value change together with it. The number of students on the missing session is determined and a new class is assigned up to that number in the session. This way, the structure of the chromosome is corrected.

## 3. Design of the Proposed Algorithm

In order to realize the proposed solution, a software consisting of three main components was developed. The architecture of the software is given in Fig 9. These components are relational database, assignment engine and user interface. The relational database is used to store data for all elements involved in the test. The resources to be used before scheduling are constrained by the database. After the exam schedule, session, course, student, appointment assignments are also kept in the database. The user interface consists of two main parts: the registration screen and the scheduling screen. The registration screen is prepared for entering and processing the constraints with the necessary information for the examination organization. The scheduling screen is the screen where the scheduling process is performed, in which resources and constraints are determined for an examination organization.

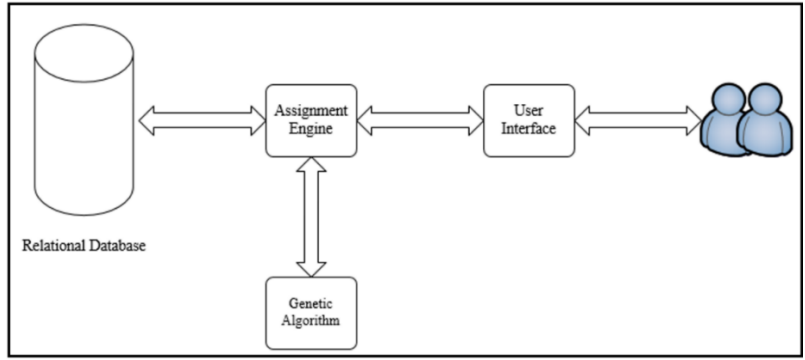

Figure 9. Software architecture

At the center of the software is the assignment engine. The assignment engine starts to work with the trigger of the user. First, it draws on the resources, constraints, courses and student data defined for the exam. It generates the initial population using captured data and operates the recommended genetic algorithm. The results obtained by the genetic algorithm are reflected in the user interface.

The tables in the database, the fields in the table and what they are used for are as follows: The examination table consists of examination, institutional fields. Examination information is entered for the procedures for the creation of examination charts. Exam information is the name of the semester or general exam to be held. The fall semester final exams or open education exams are examples. The Sessions table consists of session_no, session_name fields. After the maximum session length is determined, the number of sessions to be tested is determined and recorded according to the total number of courses.

The department table consists of department_code, department_name fields. Multiple exams can be carried out in the exam to be realized. This is why department information is entered here. Classes table consists of class_code, building_name, class_name, quota, selection fields. The classes to be used in the exams of the students are entered here. The courses table are composed of the code, course_code, course_name, credit, department_code, common_course, exam_time, selection fields. All the Courses to be examined are stored in this table. The students table consists of the fields no, name, surname, department_code. The information of the students participating in the examination is here. Student_courses table consists of no, student_no, course_code fields. This table is the table with the most data. The students and the courses they have taken are kept on this table.

As mentioned above, two separate genetic algorithms have been used to solve the problem. The best solution for size of initial populations was searched for value as soon as possible. These studies are reflected in the graph and are given in Figure 10.

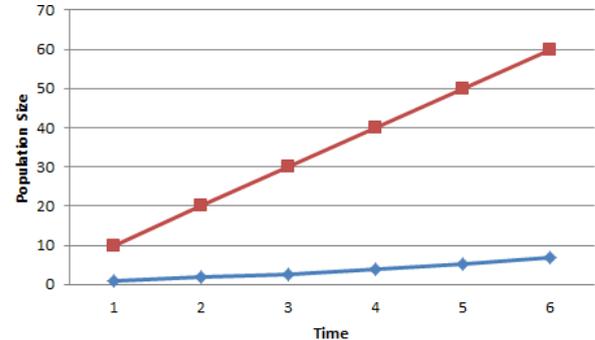

Figure 10. Population size determination

First, the solution time is set to be the population size 10. These times were determined within the other population sizes, respectively. Subsequent periods were divided into the minimum resolution period and the results were reflected. As shown in Fig. 10, the size of the population that achieved the best solution in the shortest time was determined as 10. In the genetic algorithms used, the crossover rate is 0.7 and the mutation rate is 0.05. Chromosome structures and fitness functions are different in algorithms. However, the same methods have been applied in the sub-processes of the genetic algorithm. The flow chart of the method used is given in Fig. 11.

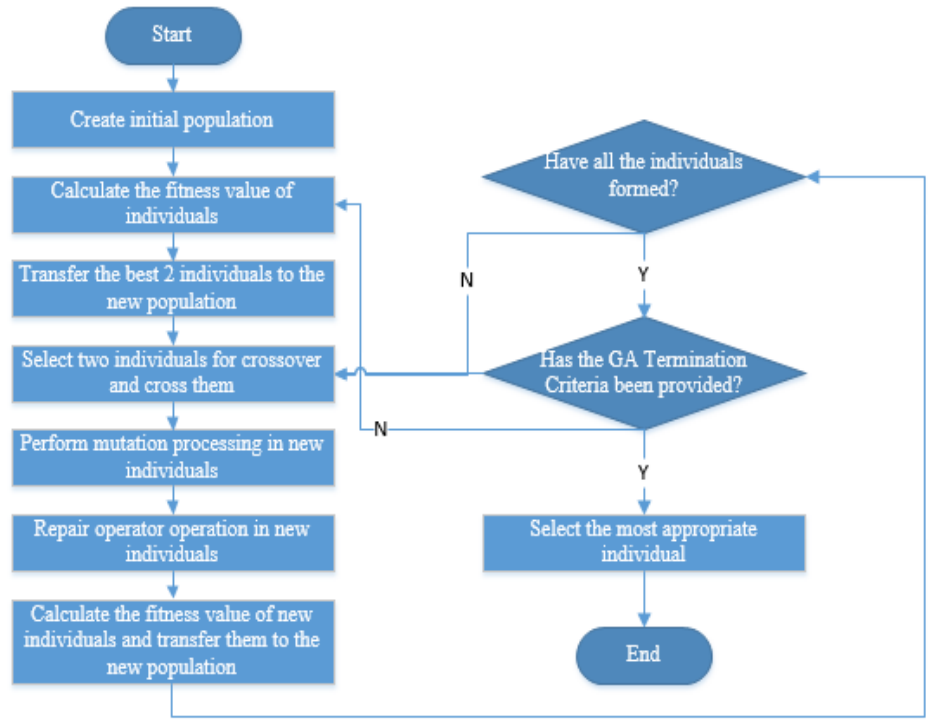

Figure 11. Flow diagram of proposed GA

The main issue in the solution of a general-purpose examination organization is that resources and constraints are variable. The resources allocated for the examination organization and the identified constraints are different for each institution for each exam period. With the developed software, institutional resources and constraints are defined in advance. Before each examination organization, it is stated that the registered resources and constraints are active and the active one is passive. Thus, developed softwarecan be used for different central exams.

## 4. Simulation Results

A sample scenario has been prepared to test the proposed genetic algorithm. A sample scenario was prepared for the visa and final sessions organized by Gazi University with 3 departments with Distance Education program. The sections of the scenario, the courses, the number of the students who took the courses and the examination periods of the courses are given in Table 1.2.3 and the information about building, class, quota, supervisor needs is given in Table 4.

Table 1. Coursesof computer programming department

| Course Code | Credit | Number of Students | Exam duration (Min.) |
|---|---|---|---|
| MATH-101 | 4 | 125 | 30 |
| FL-101 | 4 | 120 | 30 |
| TUR-101 | 2 | 115 | 30 |
| HIST-101 | 2 | 118 | 30 |
| CP-101 | 4 | 105 | 30 |
| CP-103 | 4 | 112 | 30 |
| CP-105 | 4 | 105 | 30 |

Table 2. Courses of business administration department

| Course Code | Credit | Number of Students | Exam duration (Min.) |
|---|---|---|---|
| MATH-101 | 4 | 140 | 30 |
| FL-101 | 4 | 136 | 30 |
| TUR-101 | 2 | 134 | 30 |
| HIST-101 | 2 | 130 | 30 |
| BA-101 | 4 | 133 | 30 |
| BA-105 | 4 | 132 | 30 |
| BA-107 | 2 | 130 | 30 |
| BA-109 | 2 | 128 | 30 |
| BA-103 | 2 | 125 | 30 |
| BA-111 | 2 | 128 | 30 |

Table 3. Courses of the accounting and taxation department

| Course Code | Credit | Number of Students | Exam duration (Min.) |
|---|---|---|---|
| MATH-101 | 4 | 162 | 30 |
| FL-101 | 4 | 165 | 30 |
| TUR-101 | 2 | 160 | 30 |
| HIST-101 | 2 | 158 | 30 |
| AT-101 | 4 | 160 | 30 |
| AT-103 | 2 | 152 | 30 |
| AT-105 | 4 | 162 | 30 |
| AT-107 | 2 | 155 | 30 |
| AT-109 | 2 | 158 | 30 |
| AT-113 | 2 | 160 | 30 |

Table 4. Building, class, quota, supervisor needs information

| No | Building Name | Class Name | Quota |
|---|---|---|---|
| S1 | Building_1 | Anfi_1 | 135 |
| S2 | Building_1 | Anfi_2 | 154 |
| S3 | Building_1 | Class_1 | 35 |
| S4 | Building_1 | Class_2 | 24 |
| S5 | Building_1 | Class_3 | 15 |
| S6 | Building_2 | Anfi_1 | 145 |
| S7 | Building_2 | Anfi_2 | 120 |
| S8 | Building_2 | Class_1 | 30 |
| S9 | Building_2 | Class_2 | 25 |
| S10 | Building_2 | Class_3 | 25 |
| S11 | Building_3 | Anfi_1 | 125 |
| S12 | Building_3 | Anfi_2 | 120 |
| S13 | Building_3 | Class_1 | 24 |
| S14 | Building_3 | Class_2 | 20 |
| S15 | Building_3 | Class_3 | 30 |

Note: While 3 supervisors are needed in the anfi, 1 supervisor is sufficient in the classroom.

The maximum duration of the session was 150 minutes in the scenario. As shown in Table 1.2.3, there are 27 courses in total of three departments. MATH-101, FL-101, TUR-101 and HIST-101 are common courses. There are 19 courses on a singular basis when common courses are excluded. When the courses are unified, the data in Table 5 are obtained

Table 5. Unified courses

| Code | Course Code | Credit | Number of Students | Exam duration (Min.) |
|---|---|---|---|---|
| D1 | MATH-101 | 4 | 427 | 30 |
| D2 | FL-101 | 4 | 421 | 30 |
| D3 | TUR-101 | 2 | 409 | 30 |
| D4 | HIST-101 | 2 | 406 | 30 |
| D5 | CP-101 | 4 | 105 | 30 |
| D6 | CP-103 | 4 | 112 | 30 |
| D7 | CP-105 | 4 | 105 | 30 |
| D8 | BA-101 | 4 | 133 | 30 |
| D9 | BA-105 | 4 | 132 | 30 |
| D10 | BA-107 | 2 | 130 | 30 |
| D11 | BA-109 | 2 | 128 | 30 |
| D12 | BA-103 | 2 | 125 | 30 |
| D13 | BA-111 | 2 | 128 | 30 |
| D14 | AT 101 | 4 | 160 | 30 |
| D15 | AT 103 | 2 | 152 | 30 |
| D16 | AT 105 | 4 | 162 | 30 |
| D17 | AT 107 | 2 | 155 | 30 |
| D18 | AT 109 | 2 | 158 | 30 |
| D19 | AT 113 | 2 | 160 | 30 |

According to Formula 1, the sum of the exam periods of the courses-570 min / maximum duration of one session 150 min = 3.8 = 4 sessions.

*Assigning courses to sessions:*

Chromosomes are obtained by running the course assignment algorithm. The sample initial population is given in Table 6. When a chromosome is considered in Table 6, the number of courses assigned to each session, the sum of the exam periods of the courses, and the number of common and different students to participate in the session are excluded. The values given in Table 7 are obtained when chromosome 5 is included in Table 6 is taken into consideration. The value of each chromosome is found with the aid of a fitness function. Table 8 shows the procedures for choosing roulette Wheel. Algorithms continue to work with crossover, mutation and repair operator operations. At the end of the algorithm, the chromosome given in Figure 12 stands out as the best individual.

Table 6. Sample initial population

| No | Chromosomes | | | | | | | | | | | | | | | | | | |
|---|---|---|---|---|---|---|---|---|---|---|---|---|---|---|---|---|---|---|---|
| 1 | D3 | D4 | D2 | D5 | D1 | D9 | D7 | D6 | D10 | D8 | D12 | D11 | D13 | D15 | D14 | D18 | D19 | D17 | D16 |
| 2 | D5 | D6 | D7 | D8 | D9 | D4 | D13 | D14 | D17 | D12 | D16 | D15 | D3 | D10 | D19 | D2 | D11 | D1 | D18 |
| 3 | D10 | D11 | D12 | D13 | D14 | D15 | D16 | D17 | D18 | D1 | D2 | D3 | D19 | D4 | D5 | D6 | D7 | D8 | D9 |
| 4 | D4 | D5 | D6 | D17 | D18 | D19 | D1 | D2 | D3 | D13 | D14 | D15 | D16 | D7 | D8 | D9 | D10 | D11 | D12 |
| 5 | D19 | D3 | D4 | D5 | D6 | D7 | D8 | D9 | D17 | D18 | D10 | D11 | D12 | D13 | D14 | D1 | D2 | D15 | D16 |
| 6 | D1 | D5 | D6 | D7 | D8 | D17 | D9 | D10 | D14 | D16 | D13 | D19 | D2 | D3 | D4 | D15 | D11 | D12 | D18 |
| 7 | D7 | D6 | D13 | D19 | D5 | D11 | D9 | D12 | D10 | D8 | D17 | D14 | D16 | D18 | D15 | D1 | D4 | D3 | D2 |
| 8 | D14 | D16 | D18 | D19 | D17 | D7 | D15 | D5 | D12 | D6 | D11 | D8 | D13 | D9 | D10 | D4 | D2 | D1 | D3 |
| 9 | D8 | D11 | D13 | D10 | D9 | D14 | D16 | D17 | D18 | D19 | D5 | D2 | D6 | D1 | D7 | D12 | D3 | D4 | D15 |
| 10 | D6 | D7 | D12 | D4 | D15 | D9 | D10 | D13 | D8 | D11 | D14 | D16 | D17 | D19 | D18 | D1 | D2 | D3 | D5 |

Table 7. Values related to chromosome 5

| Sessions | Assigned Courses | Exam Times Total (Min) | Total Number of Students | Number of Common Students | Number of Different Students |
|---|---|---|---|---|---|
| Session 1 | D19, D3, D4, D5, D6 | 150 | 409 | 0 | 409 |
| Session 2 | D7, D8, D9, D17, D18 | 150 | 396 | 0 | 396 |
| Session 3 | D10, D11, D12, D13, D14 | 150 | 290 | 0 | 290 |
| Session 4 | D1, D2, D15, D16 | 120 | 427 | 152 | 275 |

Table 8. Selection process

| No | Number of Common Students (CS) | Number of Different Students (DS) | (CS-DS) | =(CS-DS) – min (CS-DS) = (CS-DS) + 1487 | =ROUND(CS-DS + 1487 / Total;2)*100 | |
|---|---|---|---|---|---|---|
| 1 | 260 | 862 | -602 | 885 | 11 | 11 |
| 2 | 0 | 1487 | -1487 | 0 | 0 | 11 |
| 3 | 152 | 1231 | -1079 | 408 | 5 | 16 |
| 4 | 125 | 1240 | -1115 | 372 | 5 | 21 |
| 5 | 152 | 1370 | -1218 | 269 | 3 | 24 |
| 6 | 0 | 1428 | -1428 | 59 | 1 | 25 |
| 7 | 683 | 439 | 244 | 1731 | 22 | 47 |
| 8 | 689 | 422 | 267 | 1754 | 23 | 70 |
| 9 | 388 | 743 | -355 | 1132 | 15 | 85 |
| 10 | 388 | 740 | -352 | 1135 | 15 | 100 |
| | | | | 7745 | 100 | |

| D9 | D13 | D11 | D8 | D10 | D19 | D16 | D14 | D17 | D18 | D5 | D15 | D7 | D12 | D6 | D1 | D4 | D2 | D3 |

Figure 12. Best Individual

After this process is complete, the number of students who will participate in the test for each session and session for the next transaction is calculated as given in Table 9.

Table 9. Total number of students

| Sessions | Total Number of Students |
|---|---|
| Session 1 | 133 |
| Session 2 | 162 |
| Session 3 | 389 |
| Session 4 | 427 |

After the courses and student numbers of the session are determined, assignment to the classes is carried out. Chromosomes are obtained by running a class assignment algorithm. The sample initial population is given in Table 10.

Assigning students in session to classes:

When a chromosome is considered in Table 9, the number of classes to be used for each session, the number of vacancies, the total supervisor requirement, and the number of differences indicating the number of different buildings used in each session are extracted. When 7 chromosome in Table 9 is handled, the values given in Table 11 are obtained.

Each chromosome has value, with the help of a fitness function. Then the operations for selecting the roulette wheel are given in Table 12.

Table 10. Sample initial population

| No | Chromosomes | | | | | | | | | | | | | | | | | |
|---|---|---|---|---|---|---|---|---|---|---|---|---|---|---|---|---|---|---|
| 1 | S2 | S1 | S3 | S2 | S1 | S6 | S1 | S2 | S3 | S4 | S5 | S6 | | | | | | |
| 2 | S1 | S2 | S5 | S7 | S6 | S11 | S8 | S3 | S4 | S5 | S14 | S15 | S9 | S10 | S11 | S12 | | |
| 3 | S3 | S5 | S10 | S14 | S9 | S15 | S2 | S8 | S2 | S7 | S12 | S9 | S5 | S6 | S7 | S12 | S15 | |
| 4 | S7 | S8 | S1 | S13 | S14 | S1 | S3 | S5 | S8 | S11 | S13 | S15 | S4 | S8 | S9 | S10 | S11 | S2 | S14 | S3 |
| 5 | S10 | S12 | S4 | S6 | S3 | S7 | S11 | S12 | S15 | S9 | S10 | S2 | S1 | S11 | | | | |
| 6 | S5 | S10 | S1 | S8 | S11 | S14 | S2 | S3 | S5 | S8 | S10 | S11 | S14 | S4 | S12 | S13 | S14 | S7 | S2 |
| 7 | S1 | S3 | S4 | S7 | S6 | S7 | S9 | S10 | S11 | S13 | S7 | S8 | S9 | S11 | S12 | | | |
| 8 | S12 | S15 | S9 | S10 | S13 | S14 | S15 | S1 | S1 | S2 | S8 | S13 | S14 | S15 | S9 | S12 | S13 | S14 | S6 | S7 |
| 9 | S6 | S6 | S7 | S3 | S4 | S5 | S7 | S8 | S9 | S10 | S12 | S10 | S2 | S3 | S12 | S1 | | |
| 10 | S1 | S1 | S3 | S1 | S2 | S3 | S4 | S5 | S8 | S1 | S2 | S4 | S7 | | | | | |

Table 11. Values related to chromosome 7

| No | Total Number of Students to participate in the examination | Classes to be used | The number of vacancies in the classroom | Total supervisor requirement | Number of different buildings used in the session |
|---|---|---|---|---|---|
| 1 | 133 | S1 | 2 | 3 | 1 |
| 2 | 162 | S3, S4, S7 | 17 | 5 | 2 |
| 3 | 389 | S6, S7, S9, S10, S11 | 51 | 11 | 2 |
| 4 | 427 | S13, S7, S8, S9, S11, S12 | 17 | 12 | 2 |

Table 12. Selection process

| No | The number of vacancies in the classroom (VC) | Total supervisor requirement (TS) | Number of different buildings used in the session (DB) | VC + TS+ DB*10 | Total/ (VC + TS + DB*10) | |
|---|---|---|---|---|---|---|
| 1 | 155 | 28 | 6 | 243 | 8 | 8 |
| 2 | 32 | 30 | 7 | 132 | 16 | 24 |
| 3 | 72 | 32 | 11 | 214 | 10 | 34 |
| 4 | 50 | 32 | 9 | 172 | 12 | 46 |
| 5 | 97 | 30 | 10 | 227 | 9 | 55 |
| 6 | 105 | 33 | 10 | 238 | 9 | 64 |
| 7 | 87 | 31 | 7 | 188 | 11 | 75 |
| 8 | 145 | 34 | 9 | 269 | 8 | 83 |
| 9 | 162 | 32 | 8 | 274 | 8 | 91 |
| 10 | 20 | 27 | 6 | 107 | 19 | 110 |
| | | | | 2064 | 110 | |

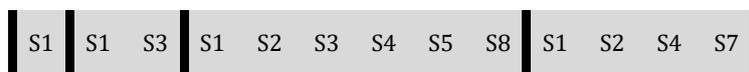

Figure 13. Best individual

Algorithms continue to work with crossover, mutation and repair operator operations. At the end of the algorithm, the chromosome shown in Figure 13 stands out as the best individual.

In this way, the application is completed. Due to the proposed two-stage genetic algorithm, the most appropriate individuals to solve are obtained as the right result.

## 5. Conclusions

Nowadays, the preparation of examination schedules in education institutions is needed very intensely. The departments becomes increasingly complicated when the number of courses and exam periods increase and the process of preparing the charts using classical methods is not possible. Test schedules prepared with classical methods cause significant disadvantages in time, labor loss and fulfillment of required limitations. Especially in the central examinations, the problem becomes more difficult with the increase of the variable number. Recently, meta-heuristic methods have been preferred in the preparation of exam schedules because of the success of schedules problem solving and soft constraints. In this study, a general purpose exam scheduling solution for educational institutions was presented. In this context, it is ensured that different institutions carry out central examination schedules to be carried out depending on different resources, targets and constraints. In the scheduling phase, a two-stage genetic algorithm is used. A sample test scenario is prepared and the results are shared. The specified objectives such as - increasing the number of participating students in sessions, minimum building use in the same session, reduce the number of supervisors using the minimum number of class-rows possible -have been achieved. GA technique is generally utilized for solving the hard problem instead of traditional methods, and can deal with difficult and easy constraints. Therefore, GA technique was used in this study, and seen that it was an effective method according to experimen results.